\documentclass[11pt,a4paper]{article}
\usepackage[hyperref]{emnlp2020}
\usepackage{array}
\usepackage{amsmath}
\usepackage{booktabs}
\usepackage{framed}
\usepackage{graphicx}
\usepackage{latexsym}
\usepackage{makecell}
\usepackage{multirow}
\usepackage{subcaption}
\usepackage{times}
\usepackage{xcolor}
\usepackage{xspace}

\usepackage{soul}
\usepackage{microtype}

\aclfinalcopy % Uncomment this line for the final submission
\newcommand\metricName{LERC\xspace}
\newcommand\dataName{MOCHA\xspace}

\title{\dataName: A Dataset for Training and Evaluating \\ Generative Reading Comprehension Metrics}

\author{
	Anthony Chen \\
	UC Irvine \\
	\texttt{\small anthony.chen@uci.edu} \\\And
	Gabriel Stanovsky\thanks{\; Work done while at the Allen Institute for AI and the University of Washington.} \\
	The Hebrew University \\
	\texttt{\small gabis@cse.huji.ac.il} \\\And
	Sameer Singh \\
	UC Irvine \\
	\texttt{\small sameer@uci.edu} \\\And
	Matt Gardner \\
	AI2 Irvine \\
	\texttt{\small mattg@allenai.org} \\
}

\newif\ifcomments
\commentstrue 
\ifcomments
    \providecommand{\sameer}[1]{\textcolor{magenta}{\textbf{sameer:} #1}}
    \providecommand{\matt}[1]{\textcolor{teal}{\bf [Matt: #1]}}
\else
    \providecommand{\sameer}[1]{}    \providecommand{\matt}[1]{}
\fi

\date{}
\begin{document}
\maketitle

\begin{abstract}
    Posing reading comprehension as a generation problem provides a great deal of flexibility, allowing for open-ended questions with few restrictions on possible answers.
    However, progress is impeded by existing generation metrics, which rely on token overlap and are agnostic to the nuances of reading comprehension.
    To address this, we introduce a benchmark for training and evaluating generative reading comprehension metrics: \textit{\textbf{MO}deling \textbf{C}orrectness with \textbf{H}uman \textbf{A}nnotations}. 
    \dataName contains 40K human judgement scores on model outputs from 6 diverse question answering datasets and an additional set of minimal pairs for evaluation.
    Using \dataName, we train a \textit{\textbf{L}earned \textbf{E}valuation metric for \textbf{R}eading \textbf{C}omprehension}, \metricName, to mimic human judgement scores.
    \metricName outperforms baseline metrics by 10 to 36 absolute Pearson points on held-out annotations.
    When we evaluate robustness on minimal pairs, \metricName achieves 80\% accuracy, outperforming baselines by 14 to 26 absolute percentage points while leaving significant room for improvement.
    \dataName presents a challenging problem for developing accurate and robust generative reading comprehension metrics.\footnote{The dataset, code, a leaderboard, and a demo are available at \url{https://allennlp.org/mocha}.}
\end{abstract}
\section{Introduction}
    Reading comprehension (RC) has seen significant progress in the last few years, with a number of question answering (QA) datasets being created \citep{Rajpurkar2016SQuAD10,Lai2017RACELR,Talmor2018CommonsenseQAAQ}.
    However, a majority of datasets are presented using a span-selection or multiple-choice (MC) format.
    Both formats are easy to evaluate, but in return, have restrictions placed on the questions that can be asked or the answers that can be returned.
    Furthermore, both formats hinge on distractor spans/choices for learning to be effective.
	Ensuring high quality distractors is a challenging task in and of itself, which can lead to models that exploit spurious correlations \citep{Jia2017AdversarialEF, Min2019CompositionalQD, Geva2019AreWM}.
	Posing RC as a generation task addresses the aforementioned issues.
	Generative RC does not require distractors, circumventing biases that could be introduced by them, and allows arbitrary questions and answers.
	
	% Figure on metric flaws
    \begin{figure}[t!]
        \begin{framed}
            \small \textbf{Passage:} \ldots Behind one door is a lady whom the king has deemed an appropriate match for the accused; behind the other is a fierce, hungry tiger. Both doors are \textbf{\textcolor{purple}{heavily soundproofed to prevent the accused from hearing what is behind each one}}\ldots
            \vskip 2mm
            {\small
            \begin{tabular}{@{}lp{50mm}}
            \textbf{Question:} & What feature do the doors have?
            \end{tabular}
            \begin{tabular}{@{}lp{45mm}}
            \textbf{Reference:} & soundproofed \\
            \textbf{Candidate:} & They are \textbf{\textcolor{purple}{heavily soundproofed to prevent the accused from hearing what's behind each one}}.
            \end{tabular}
            }
            \vskip 2mm
            \small \textbf{Human Judgement:} 5 out of 5\\
            \small \textbf{LERC:} 4.98 out of 5 \vspace{1.5mm}\\
            \small \textbf{BLEU-1:} 0.07 \\
            \small \textbf{ROUGE-L:} 0.15 \\
            \small \textbf{METEOR:} 0.17
        \end{framed}
        
        \caption{Generative reading comprehension example. Properly scoring the candidate requires access to the passage. Current metrics, such as BLEU, ROUGE and METEOR, are agnostic to the end-task while \metricName is trained with the passage and question as input. As a result, \metricName assigns a score that better reflects human judgement.}
        \label{fig:introduction:metric-flaws}
    \end{figure}  

	Unfortunately, existing metrics for evaluating text generation come with significant shortcomings.
	Many metrics score \textit{n}-gram overlap, and it is well established that using token overlap as a measure of similarity has drawbacks \citep{Chen2019EvaluatingQA,Edunov2019OnTE,Wang2020AskingAA}.
	Current metrics also only consider the reference and are agnostic to the end-task being evaluated.
	Fig. \ref{fig:introduction:metric-flaws} demonstrates that this is problematic for generative RC because scoring a candidate may require a metric to also consider the passage and the question.
	Without cheap and reliable evaluation, progress in generative reading comprehension has been extremely slow.
	
	To address the need for better evaluation metrics tailored to reading comprehension, we present a dataset called \dataName, aimed at developing \textit{learned} metrics that \textbf{MO}del the \textbf{C}orrectness of candidates using \textbf{H}uman \textbf{A}nnotation scores.
	\dataName contains human judgement scores on 40K candidates, an order of magnitude larger than prior work~\citep{Chen2019EvaluatingQA}.
	The candidates come from six diverse QA datasets which test a wide range of RC phenomena such as commonsense reasoning and understanding narrative over movie scripts.
	After collecting all annotations, we follow work on creating more robust evaluation sets~\citep{Kaushik2020LearningTD, Gardner2020EvaluatingNM} and augment the test set of \dataName by manually writing a small set of minimal pairs (Table \ref{tab:dataset:minimal-pairs}).
	The set of minimal pairs serve as a harder evaluation set for probing metric robustness.
	
	Using \dataName, we train a \textbf{L}earned \textbf{M}etric for \textbf{R}eading \textbf{C}omprehension which we abbreviate as LERC.
	We compare \metricName against two sets of baselines: (1) existing metrics such as METEOR~\citep{Banerjee2005METEORAA} and BERTScore~\citep{Zhang2019BERTScoreET}; and (2) a sentence similarity model trained on STS-B~\citep{Cer2017SemEval2017T1}.
	To ensure fair comparison, we evaluate \metricName in an out-of-dataset setting: \metricName is trained on all datasets \textit{except} the one it is being evaluated on.
	On the test set, \metricName outperforms baselines by as much as 36 Pearson correlation points and on the minimal pairs set, by as much as 26 accuracy points.
    Error analysis and minimal pair results indicate that there is substantial room to improve the robustness of \metricName and its sensitivity to different linguistic phenomena.
    We hope that \dataName and \metricName enables a continual cycle of generative RC model and dataset developments that will enable easier collection of more diverse and useful candidates, allowing better learned metrics to be trained.
\begin{table}[h!t]
        \small
        \begin{center}
        \begin{tabular}{p{2.3 in} c}
        	\toprule
            \bf Instance & \bf Score\\
            \midrule
            \textbf{Passage:} With the aid of his daughter, Abigail, Barabas recovers his former assets. Barabas then uses his daughter's beauty to embitter Lodowick and Mathias against each other. \newline 
            \textbf{Q:} Why did Lodowick and Mathias fight? \newline 
            \textbf{Ref:} Over the affection of Abigail \newline 
            \textbf{Cand:} They fight over Barabas's daughter.
            & 5 \\
            \addlinespace
            
            \textbf{Passage:} Miss Moppet ties a duster about her head and sits before the fire. The mouse thinks she looks very ill and comes down the bell-pull.  \newline 
            \textbf{Q:} What does the mouse think when she sees the duster on Miss Moppet's head?\newline 
            \textbf{Ref:} that Miss Moppet is ill \newline
            \textbf{Cand:} Miss Moppet thinks it is ill and is trying to sniff him. \newline
            & 2 \\
        
            \textbf{Passage:} Robin took a very long time to clean the windows of her house. \newline 
            \textbf{Q:} How would you describe Robin? \newline 
            \textbf{Ref:} a neat freak \newline 
            \textbf{Cand:} a clean person \newline
            & 5  \\
         
            \textbf{Passage:} The strangest thing that has happened was when they were singing the Chinese National Anthem she was standing in front of the TV swaying and singing. \newline 
            \textbf{Q:} What is probably true about this story? \newline 
            \textbf{Ref:} They are watching the Olympics \newline
            \textbf{Cand:} The Olympics are watching \newline
            & 2 \\
            \bottomrule
        \end{tabular}
        \end{center}
            \caption{Example instances with human judgement scores from \dataName highlighting the diverse phenomenon that an evaluation metric needs to handle. These phenomenon include resolving coreference, dealing with factual correctness, understanding paraphrases, and understanding semantic roles.}
        \label{tab:dataset:examples}
    \end{table}

\section{A Description of \dataName}
    % FIGURE FOR ANNOTATION INTERFACE
    \begin{figure*}[h!t!]
        \includegraphics[width=\linewidth]{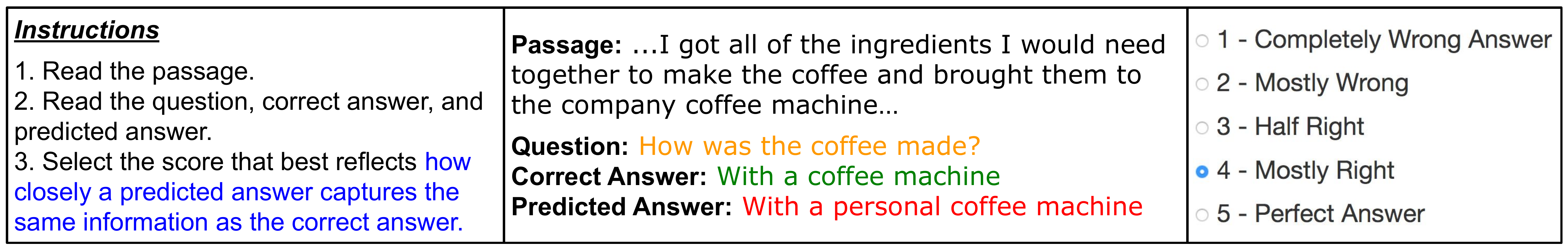}
        \caption{A compressed version of the Mechanical Turk interface for evaluating answer correctness. Workers were asked to score (1 to 5) how similar a candidate is to a reference \textbf{using} the passage and the question.}
        \label{fig:dataset:interface}
    \end{figure*}
    
    Reading comprehension is the task of probing how well systems can understand passages of text. Framing reading comprehension as a generation problem provides a great deal of flexibility, but introduces the challenging problem of evaluation.
    These challenges are further amplified when applied to generative reading comprehension, where the introduction of a passage and a question can add to the complexity of evaluation (Table \ref{tab:dataset:examples}).
    To handle this challenge, we propose to \textit{train} a generative reading comprehension metric.
    This first requires a large set of human judgement scores to be gathered.
    
	In this section, we present \dataName, a dataset that pairs reading comprehension instances, which consists of a passage, question, and reference, with candidates and human judgement scores.
	We  describe the process of gathering candidates, collecting human judgement scores, and creating minimal pairs for evaluation. 

	\subsection{Datasets}
	    Candidates in \dataName come from 6 constituent QA datasets that are diverse in their domains and answer types.
	    This ensures that training and evaluation with \dataName does not overfit to the characteristics of any constituent dataset.
	    
		\paragraph{\fontfamily{cmss}\selectfont NarrativeQA} \citep{Kocisk2017TheNR}
		tests reasoning about events, entities, and their relations on movie scripts and book summaries.

		\paragraph{\fontfamily{cmss}\selectfont MCScript} \citep{Ostermann2018MCScriptAN}
		tests reasoning on stories written for a child-level reader. 

		\paragraph{\fontfamily{cmss}\selectfont CosmosQA} \citep{Huang2019CosmosQM}
		tests commonsense reasoning on blogs describing everyday events.

		\paragraph{\fontfamily{cmss}\selectfont SocialIQA} \citep{Sap2019SocialIC}
		tests social reasoning with passages constructed from a knowledge base.

		\paragraph{\fontfamily{cmss}\selectfont DROP} \citep{Dua2019DROPAR}
		tests predicate argument structure and numerical reasoning on Wikipedia articles concerning American football games, census results, and history.

		\paragraph{\fontfamily{cmss}\selectfont Quoref} \citep{Dasigi2019QuorefAR}
		tests coreferential reasoning on Wikipedia articles.\\
        
        NarrativeQA was created as a generative RC dataset.
	    CosmosQA, MCScript, and SocialIQA were created as MC datasets which we re-purpose as generative datasets by using the correct choice as the reference.
	    Our motivation for doing this is that the number of generative QA datasets is quite small, which we attribute to the quality of  evaluation metrics.
	    
	    The main focus of this work is in developing and evaluating metrics for generative RC. However, we wanted to see whether a learned metric could do well on span-selection datasets.
	    We collected candidates on two span-based datasets, DROP and Quoref, to test this.
	    
    \subsection{Collecting Candidates}
	    Candidates on all four generative datasets are generated using backtranslation \citep{Sennrich2016ImprovingNM} and using a fine-tuned GPT-2 model \citep{Radford2019LanguageMA}.
	    We also generate candidates for NarrativeQA and MCScript using a trained MHPG model \citep{Bauer2018CommonsenseFG}.
	    We tried using MHPG for CosmosQA and SocialIQA but candidates were of poor quality.
	    Unique to NarrativeQA, each question has two references. 
	    We treat the second reference as a candidate to be annotated if it has low \textit{n}-gram overlap with the first reference.
	    We use a span-selection BERT-based model to generate candidates for Quoref and NAQANET~\cite{Dua2019DROPAR} and NABERT\footnote{https://github.com/raylin1000/drop-bert} models for DROP.

		Models are trained on the training sets of each constituent dataset and candidates are produced on instances from the validation set (and test set if available).
		We filtered out candidates that exactly matched the reference.
		We also filtered out instances in DROP where the reference and the candidate are both numbers.\footnote{From our inspection, if the reference and candidate are both numbers that are not equal, the candidate is always wrong.}
        
        In total, \dataName contains 40K candidates, large enough for training a learned metric as well as for evaluating current and future metrics.
    
    % TABLE HARC STATISTICS
    \begin{table*}[t!h!]
    	\centering
    	\resizebox{16cm}{!}{
    	\begin{tabular}{lccccccccccccc}
    	\toprule 
    	\multirow{2}{*}{\bf  Dataset} & \multirow{2}{*}{\bf  \thead{Avg \\Pass. Len}} & \multirow{2}{*}{\bf \thead{Avg \\ Ques. Len}} & \multirow{2}{*}{\bf \thead{Avg \\ Ref. Len}} &  \multirow{2}{*}{\bf \thead{Avg \\ Cand. Len}} & \multicolumn{3}{c}{\bf  \# Passages} & \multicolumn{3}{c}{\bf  \# Ques./Ref. Pairs} & \multicolumn{3}{c}{\bf  \# Candidates} \\
        \cmidrule(lr){6-8}
        \cmidrule(lr){9-11}
        \cmidrule(lr){12-14}
    	& & & & & Train & Dev & Test & Train & Dev & Test & Train & Dev & Test \\\midrule
    		NarrativeQA & 333.0 & 9.6 & 5.8 & 5.9 & 85 & 11 & 18  & 2249 & 277 & 500 & 7471 & 890  & 1707 \\ 
    		MCScript 	& 197.1 & 7.8 & 4.3 & 4.1 & 462 & 61 & 93 & 2940 & 390 & 583 & 7210 & 978 & 1409 \\
    		CosmosQA 	& 72.8 & 10.8 & 7.5 & 8.8 & 1064 & 142 & 212 & 1139 & 156 & 226 & 5033 & 683 & 1017 \\
    		SocialIQA 	& 15.7 & 7.2  & 3.9 & 3.9 & 3075 & 414 & 611 & 3075 & 414 & 611 & 7409 & 1017 & 1527 \\
    		DROP 		& 213.4 & 11.6 & 3.6 & 5.1 & 80 & 10 & 17 & 542 & 76 & 117 & 687 & 97 & 152 \\
    		Quoref 		& 324.0 & 15.8 & 2.3 & 8.2 & 184 & 24 & 38 & 1098 & 123 & 180 & 3259 & 344 & 509\\ 
    		\midrule
    		Total       &   &   &   &   & 4950 & 662 & 989 & 11043  & 1436 & 2217 & 31069 & 4009  & 6321 \\
    	\bottomrule
    	\end{tabular}
    	}
    	\caption{Statistics for the human judgements per constituent dataset in \dataName.}
    	\label{table:dataset:HARC-stats}
    \end{table*}    

	\subsection{Annotation Procedure}
    	Annotations are collected with Mechanical Turk using the interface in Fig. \ref{fig:dataset:interface}.
        Workers are asked to score candidate answers on an ordinal scale from 1 to 5.
    	We start by collecting a single annotation per candidate.
    	Following this, candidates are split into training, validation, and test sets such that all candidates from a passage are contained within a dataset split.
        For instances in our validation and test sets, we collect one additional human judgement score per candidate for span-based datasets, and two additional human judgement scores per candidate for generative datasets.
        Multiple annotations for a given candidate are averaged to form a gold annotation.
	    More details such as payout and qualification testing are provided in Appendix \ref{sec:appendix_mechanical_turk}.
	    
	    We calculated inter-annotator agreement using Krippendorff's Alpha-Reliability~\citep{Krippendorff2011ComputingKA} on the validation set of all 6 constituent datasets. We choose this metric because it applies to our setting, where there are multiple annotators per instance, and the annotators vary between instances. Agreement on our 6 datasets range from 0.71 to 0.92 (average = 0.82), indicating strong agreement. 

    % FIGURE FOR SCORE DISTRIBUTIONS
    \begin{figure}[t]
        \centering
    	\includegraphics[width=0.8\linewidth]{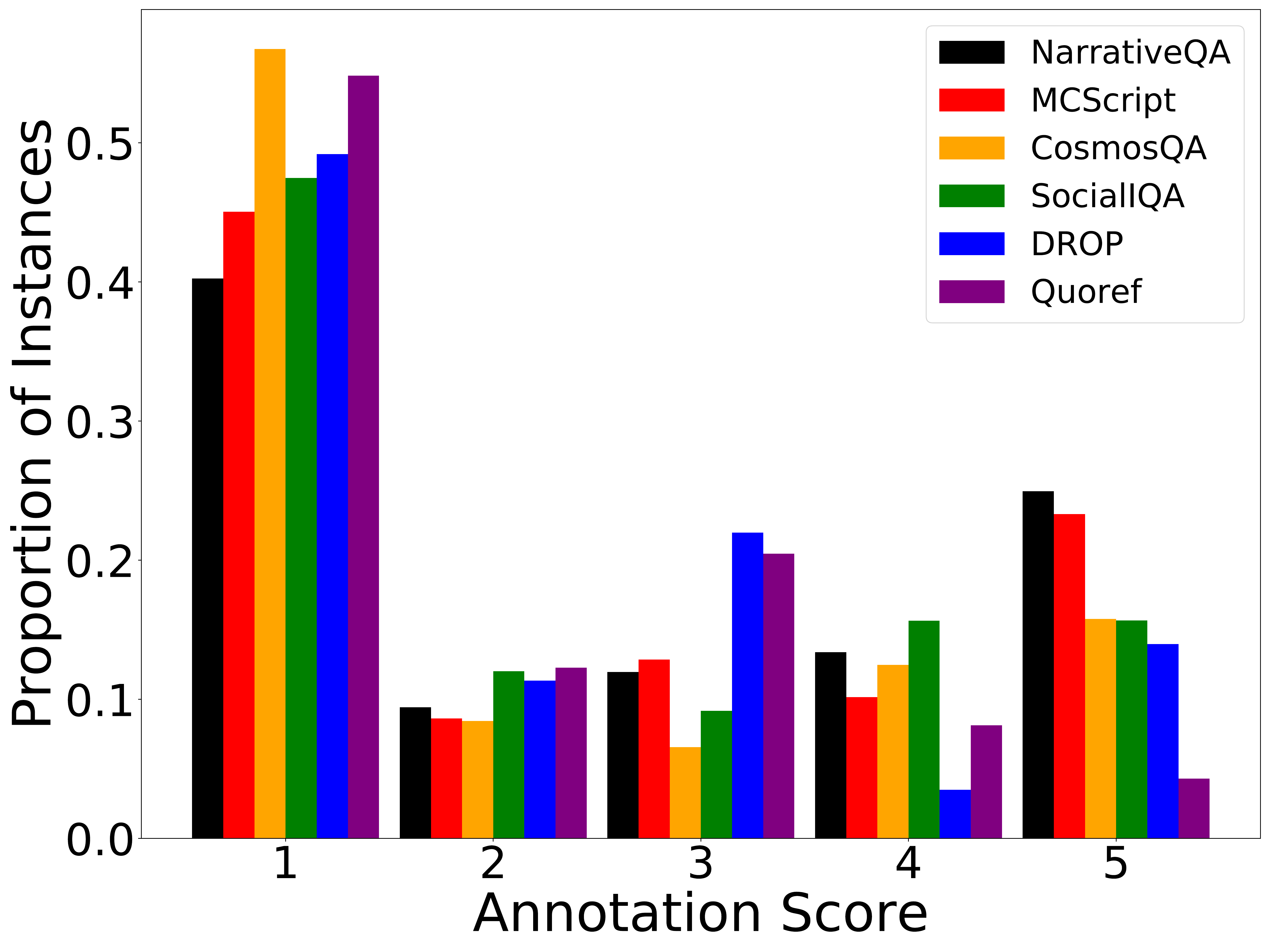}  
        \caption{Human judgement score distribution on the training set of \dataName, divided into the 6 constituent datasets. The distribution of scores is right-skewed because we did not annotate candidates that exactly matched a reference.}
    	\label{fig:dataset:all-scores}
    \end{figure}
    
    \begin{figure}[t]
        \centering
    	\includegraphics[width=0.8\linewidth]{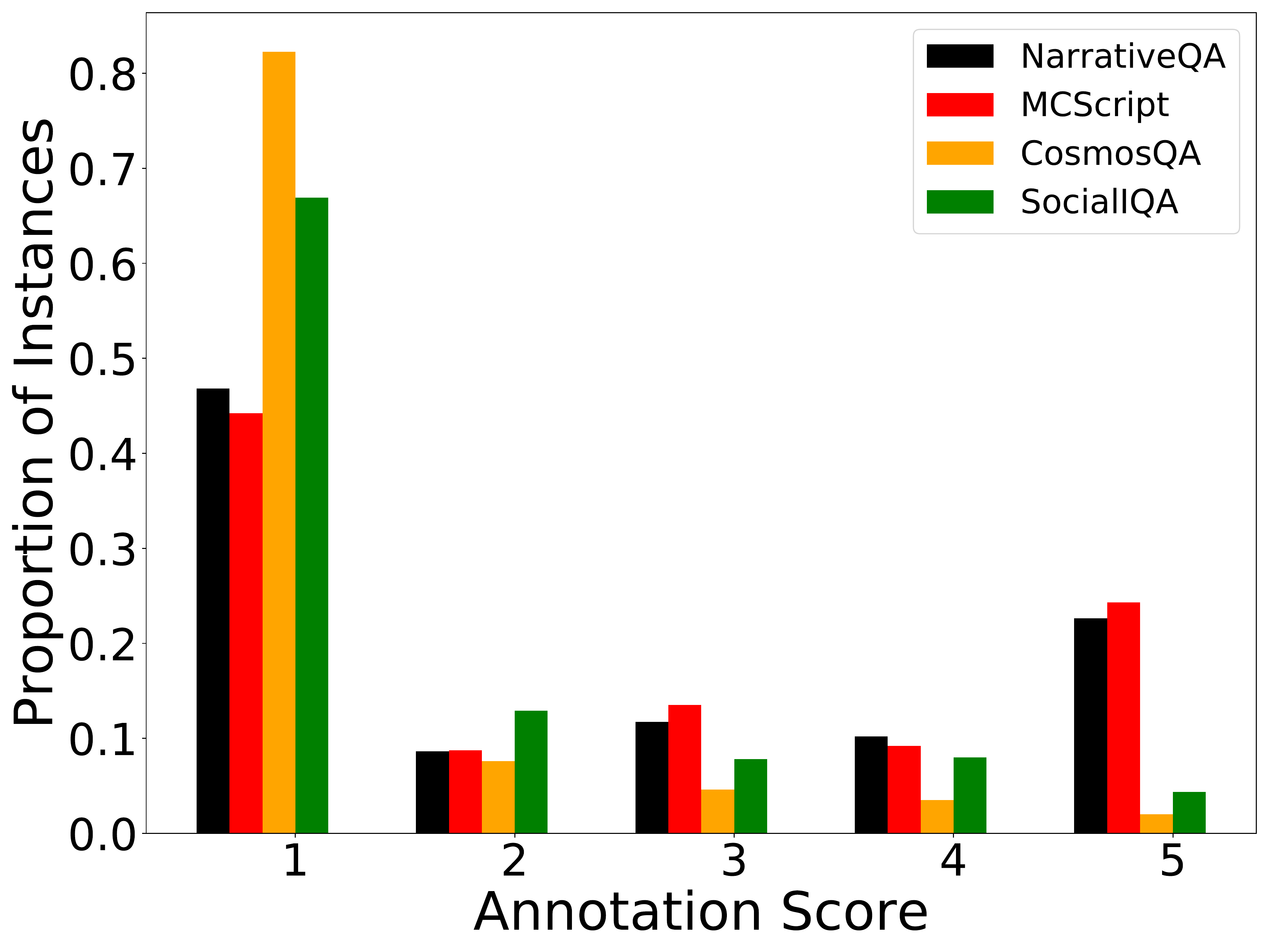}  
        \caption{Score distribution on candidates from GPT-2. GPT-2 produces a \textit{very} skewed score distribution for CosmosQA and SocialIQA, highlighting the difficulty of generative RC on commonsense questions.}
    	\label{fig:dataset:gpt2-scores}
    \end{figure}    
    
    \subsection{Statistics for \dataName}
        Statistics of instances and dataset splits in \dataName are provided in Table \ref{table:dataset:HARC-stats}.
        The number of unique passages varies considerably across datasets.
        NarrativeQA, which has the longest passages, has few unique passages, while SocialIQA has a unique passage for each question/reference pair.
        The number of candidates also varies across datasets.
        The most pronounced outlier is DROP, where we collected a tenth of the candidates compared to the other datasets. 
        This is because we filtered out instances when both the candidate and reference were numbers, leaving much fewer candidates to annotate.
        The number of candidates outnumbers the question/reference pairs because for each pair, we generated multiple candidates using different generation sources (e.g. backtranslation, different model outputs).
        
        Fig. \ref{fig:dataset:all-scores} provides the annotation score distribution on the training set of \dataName.
        Score distributions are right-skewed because we did not collect annotations when the reference exactly matched the candidate.
        The right-skew is most pronounced for Quoref because the number of ways a candidate can get a perfect score while not matching the reference is limited in a span extraction format.

    % TABLE FOR MINIMAL PAIRS
    \begin{table*}[t!]
        \centering
        \small 
        \begin{tabular}{p{.7in} p{2.75in} p{2.1in}}
          \toprule \textbf{Phenomenon}   &  \textbf{Original Instance} & \textbf{Minimal Pairs}\\ \midrule
          
          Coreference  & 
            \textbf{Passage:} Norman is the supposed son of Frenchman de Vac \ldots As de Vac dies, he reveals Norman is Richard, the king's son and Edward's brother, who he kidnapped. \newline 
            \textbf{Q:} Who is the Frenchman de Vac? \newline
            \textbf{Ref:} a fencing master who kidnapped Norman & 
            
            \textbf{\textcolor{blue}{Cand. 1:}} a fencing master who kidnapped Richard \textbf{\textcolor{olive}{(5)}} \vspace{1mm} \newline 
            \textbf{\textcolor{red}{Cand. 2:}} a fencing master who kidnapped Edward \textbf{\textcolor{olive}{(3)}}\\\\
          
          Hyponymy  & 
            \textbf{Passage:} With the electric rifle, Tom and friends bring down elephants, rhinoceroses, and buffalo. \newline 
            \textbf{Q:} What does Tom bring down with his rifle? \newline
            \textbf{Ref:} Rhinoceroses, buffalo, and elephants. &
            
            \textbf{\textcolor{blue}{Cand. 1:}} Animals \textbf{\textcolor{olive}{(4)}} \vspace{1mm} \newline 
            \textbf{\textcolor{red}{Cand. 2:}} Humans \textbf{\textcolor{olive}{(1)}}\\\\
          
          Negation  & 
            \textbf{Passage:} skylar told quinn's friend about a secret that quinn wanted to keep hidden. \newline 
            \textbf{Q:} What will Quinn want to do next? \newline
            \textbf{Ref:} be angry & 
            
            \textbf{\textcolor{blue}{Cand. 1:}} Quinn will be mad at Skylar \textbf{\textcolor{olive}{(5)}} \vspace{1mm} \newline 
            \textbf{\textcolor{red}{Cand. 2:}} Quinn will not be mad at Skylar \textbf{\textcolor{olive}{(1)}}\\\\
          
          Semantic Role  & 
            \textbf{Passage:} Taylor gave a raise and promotion to Kendall. \newline 
            \textbf{Q:} How would you describe Taylor? \newline
            \textbf{Ref:} As someone who appreciates what Kendall does & 
            
            \textbf{\textcolor{blue}{Cand. 1:}} Taylor appreciates Kendall \textbf{\textcolor{olive}{(5)}} \vspace{1mm} \newline 
            \textbf{\textcolor{red}{Cand. 2:}} Kendall appreciates Taylor \textbf{\textcolor{olive}{(1)}}\\\\
          
          Syntax  & 
            \textbf{Passage:} Taylor looked around in Robin's cupboards and peeked inside Robin's drawers and medicine cabinet. \newline 
            \textbf{Q:} How would you describe Taylor? \newline
            \textbf{Ref:} intrusive &
            
            \textbf{\textcolor{blue}{Cand. 1:}} I would describe Taylor as intrusive \textbf{\textcolor{olive}{(5)}} \vspace{1mm} \newline
            \textbf{\textcolor{red}{Cand. 2:}} Would I describe Taylor as intrusive \textbf{\textcolor{olive}{(3)}}\\\\
          
          Word Sense  & 
            \textbf{Passage:} Taylor got married but kept her last name. \newline 
            \textbf{Q:} How would you describe Taylor? \newline
            \textbf{Ref:} independent &
            
            \textbf{\textcolor{blue}{Cand. 1:}} individualistic \textbf{\textcolor{olive}{(5)}} \vspace{1mm} \newline 
            \textbf{\textcolor{red}{Cand. 2:}} nonpartisan \textbf{\textcolor{olive}{(1)}}\\\\
            
          Other  & 
            \textbf{Passage:} The Princess stuffs her ears with cotton and begins her journey.  \newline 
            \textbf{Q:} What does the Princess put in her ears?  \newline
            \textbf{Ref:} She puts cotton in her ears. &
            
            \textbf{\textcolor{blue}{Cand. 1:}} Her ears have cotton  \textbf{\textcolor{olive}{(4)}} \vspace{1mm} \newline 
            \textbf{\textcolor{red}{Cand. 2:}} Her ears are cotton \textbf{\textcolor{olive}{(2)}}\\
            
          \bottomrule
        \end{tabular}
        \caption{Minimal pairs categorized by the linguistic phenomena. Given a passage, question, and reference, we create two new candidates, \textcolor{blue}{$\mathbf{c_1}$} and \textcolor{red}{$\mathbf{c_2}$}, with associated human judgement scores \textcolor{olive}{$\mathbf{s_1}$} and \textcolor{olive}{$\mathbf{s_2}$}. In total, we wrote 200 minimal pairs (50 for each generative QA dataset).}
        \label{tab:dataset:minimal-pairs}
    \end{table*}
    
    \subsection{Limitations and Robust Evaluation with Minimal Pairs}
        Candidates in \dataName come from existing models, so that a metric learned on this data will be most applicable to current research. 
        However, as research in generative reading comprehension models is presently limited, the strength of these models can be low.
        Fig. \ref{fig:dataset:gpt2-scores} shows that generative QA models struggle to produce quality answers when asked about commonsense scenarios.
        The majority of 5's in CosmosQA and SocialIQA are produced via backtranslation, while GPT-2 struggles to produce ``correct'' candidates.
        This raises an issue with the evaluation; a metric can look strong when evaluated on current model outputs, but may in-fact struggle in the future when QA systems produce better answers.
        Thus, using only these candidates for evaluation could lead to overconfidence in a learned metric's capabilities.

        We take inspiration from from recent work creating more robust evaluations \citep{Kaushik2020LearningTD, Gardner2020EvaluatingNM} and augment the test set of \dataName with a small number of minimal pairs created by the authors.
    	Given a passage, question, and reference from the test set, we manually create two new candidates, $\mathbf{c_1}$ and $\mathbf{c_2}$, which form a minimal pair.
    	Accompanying $\mathbf{c_1}$ and $\mathbf{c_2}$ are human judgement scores, $\mathbf{s_1}$ and $\mathbf{s_2}$, collected using the same interface in Fig. \ref{fig:dataset:interface}.
    	The minimal pair is created so that $\mathbf{c_1}$ has a higher score (i.e. is a better answer) than $\mathbf{c_2}$.
    	Each minimal pair is designed to capture a particular linguistic phenomenon (see Table \ref{tab:dataset:minimal-pairs}).
    	Using this set of minimal pairs, we can study how often a metric prefers the better candidate.
    	We create 200 minimal pairs (50 for each generative QA dataset), which we use for evaluation \textit{separately} from the original test set.
    	
% Figure on model
\begin{figure}[t!]
    \includegraphics[width=\linewidth]{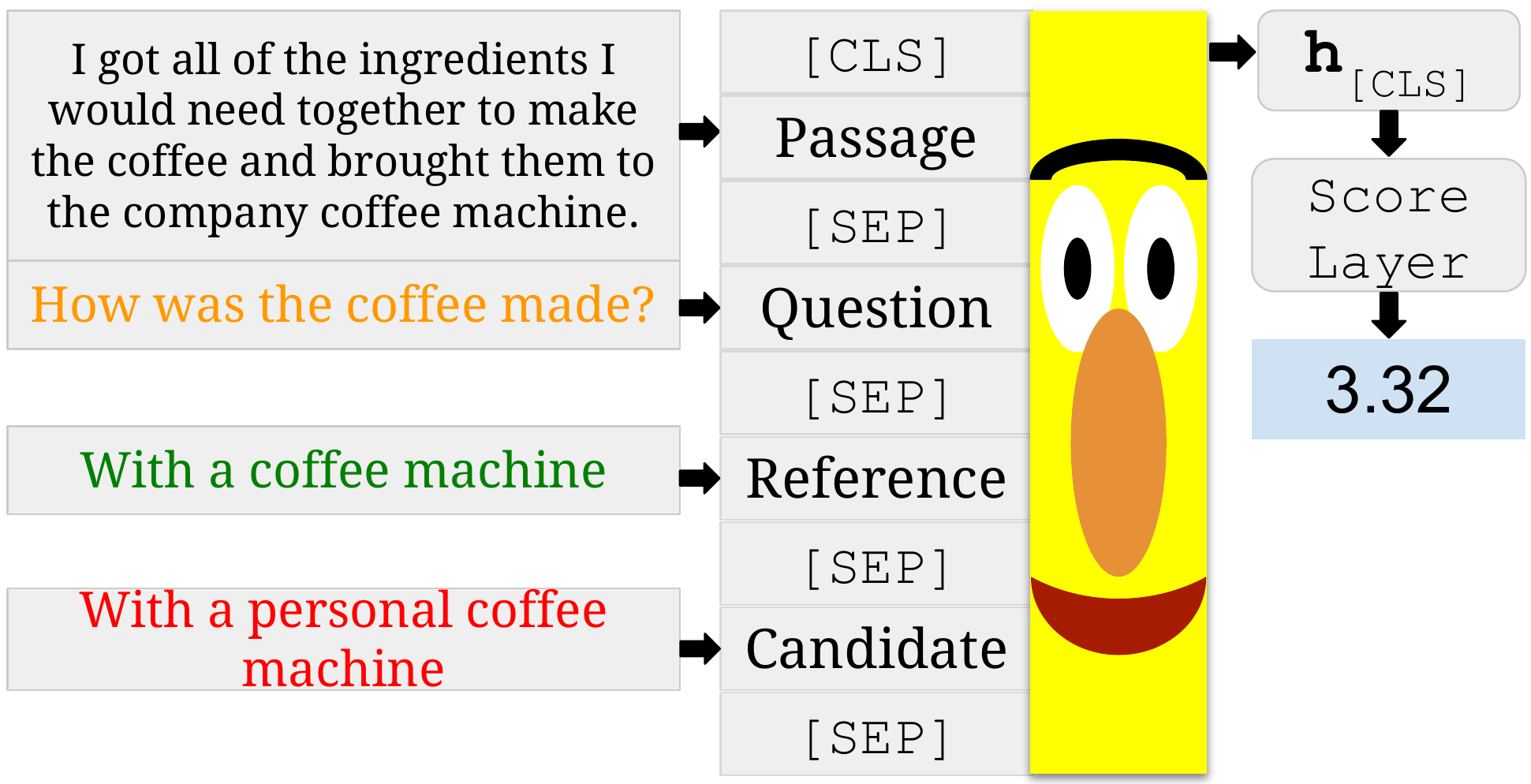}
    \caption{\metricName is a BERT model that has been fine-tuned on human judgment scores. \metricName takes as input a passage, question, reference, and candidate, and returns a score rating the "correctness" of the candidate.}
    \label{fig:model}
\end{figure} 

    \begin{table*}[t!]
        \begin{center}
        \resizebox{16cm}{!}{
        \begin{tabular}{lcccccccccccccc}
        	\toprule
            % \bf  Metric & \bf  NarrativeQA & \bf  MCScript & \bf  CosmosQA & \bf SocialIQA & \bf DROP & \bf Quoref \\
            \multirow{2}{*}{\bf  Metric} & \multicolumn{2}{c}{\bf  NarrativeQA} & \multicolumn{2}{c}{\bf  MCScript} & \multicolumn{2}{c}{\bf CosmosQA} & \multicolumn{2}{c}{\bf SocialIQA} & \multicolumn{2}{c}{\bf DROP} & \multicolumn{2}{c}{\bf Quoref}& \multicolumn{2}{c}{\bf Avg. \textit{r}} \\
            & Dev & Test & Dev & Test & Dev & Test& Dev & Test & Dev & Test & Dev & Test & Dev & Test\\ \midrule
            BLEU-1      	& 0.403 & 0.472 & 0.181 & 0.260 & 0.660 & 0.670 & 0.595 & 0.549 & 0.409 & 0.387 & 0.674 & 0.578 & 0.487 & 0.486 \\
            METEOR      	& 0.605 & 0.615 & 0.461 & 0.502 & 0.696 & 0.711 & 0.644 & 0.637 & 0.664 & 0.568 & \textbf{0.729} & 0.716 & 0.633 & 0.624 \\
            ROUGE-L 	    & 0.434 & 0.495 & 0.224 & 0.297 & 0.701 & 0.701 & 0.599 & 0.558 & 0.480 & 0.366 & 0.712 & 0.604 & 0.525 & 0.503 \\
            BERTScore   	& 0.419 & 0.534 & 0.172 & 0.194 & 0.803 & 0.779 & 0.604 & 0.584 & 0.174 & 0.328 & 0.207 & 0.286 & 0.396 & 0.450 \\
            \addlinespace \hline \addlinespace
            BERT STS-B      & 0.711 & 0.686 & 0.364 & 0.449 & 0.803 & 0.789 & 0.663 & 0.666 & 0.690 & \textbf{0.715} & 0.690 & \textbf{0.750} & 0.653 & 0.676 \\
            \addlinespace \hline \addlinespace
            \metricName     & \textbf{0.772} & \textbf{0.738} & \textbf{0.666} & \textbf{0.694} & \textbf{0.852} & \textbf{0.824} & \textbf{0.777} & \textbf{0.799} & \textbf{0.760} & 0.712 & 0.704 & 0.741 & \textbf{0.755} & \textbf{0.751} \\
            \bottomrule 
        \end{tabular}
        }
        \end{center}
        \caption{Pearson correlation to human judgement scores on the validation and test sets of \dataName. LERC results are from a model trained in an out-of-dataset fashion, averaged across three runs.} 
        \label{tab:experiments:metric-correlation}
    \end{table*}
    
\section{A Learned Metric}
    We provide details on \metricName, our learned metric.
    \metricName is initialized using BERT-base \citep{Devlin2018BERTPO} 
    We define as input a tuple consisting of a passage, $\textbf{p}$, a question, $\textbf{q}$, a reference answer, $\textbf{a}$, and a candidate answer, $\hat{\textbf{a}}$. 
    The input to BERT is structured as:
    \begin{align*}
    \text{[CLS]}\;\text{\textbf{p}}\;\text{[SEP]}\;\text{\textbf{q}} \;\text{[SEP]}\;\text{\textbf{a}}\;\text{[SEP]}\;\hat{\text{\textbf{a}}}\;\text{[SEP]}
    \end{align*}
    BERT returns a hidden state for each input token. We use the first hidden state $\textbf{h}_{[CLS]}$, as the pooled representation of the input.

    \subsection{Fine-Tuning with Human Judgements}
        Our goal is to train BERT to mimic the human judgements given a set of input tuples, $\{(\textbf{p}, \textbf{q}, \textbf{a}, \hat{\textbf{a}})\}_{i=1}^n$, and a set of human judgment scores, $\{y\}_{i=1}^n$, 
        We apply a regression layer on top of our pooled representation (Fig. \ref{fig:model}) and train with a MSE loss.
        \begin{gather*}
        \hat{y_i} = \textbf{W}\:\textbf{h}_{i\:[CLS]} \\
        \text{loss}_i = (y_i - \hat{y}_i)^2
        \end{gather*}
    
    \subsection{Pre-Training the Learned Metric}
        Learning the interactions between the input components can be difficult with only human judgement fine-tuning.
        To overcome this, we \textit{pre-train} on four multiple-choice QA datasets: BoolQ \citep{Clark2019BoolQET}, MCTest \citep{Richardson2013MCTestAC}, RACE \citep{Lai2017RACELR}, and MultiRC \citep{Khashabi2018LookingBT}.
        We use the same input structure as fine-tuning, but the reference and candidate are replaced by two answer choices, $\mathbf{a_1}$ and $\mathbf{a_2}$:
        \begin{align*}
            \text{[CLS]}\;\text{\textbf{p}}\;\text{[SEP]}\;\text{\textbf{q}} \;\text{[SEP]}\;\mathbf{a_1}\;\text{[SEP]}\;\mathbf{a_2}\;\text{[SEP]}
        \end{align*}
        We pre-train BERT via 3-way classification to predict whether: $\mathbf{a_1}$ is the correct answer, $\mathbf{a_2}$ is the correct answer, or $\mathbf{a_1}$ and $\mathbf{a_2}$ are both correct. 
        MultiRC has multiple correct answers per question and we create additional instances where both $\mathbf{a_1}$ and $\mathbf{a_2}$ are correct by duplicating the correct answer for all three datasets.
\section{Experiments}
    \textbf{Training LERC:}
    We use the PyTorch \citep{Paszke2019PyTorchAI}, HuggingFace Transformers \citep{Wolf2019HuggingFacesTS}, and AllenNLP \citep{Gardner2017ADS} libraries to implement \metricName.
    We pre-train \metricName before fine-tuning on \dataName.
    We evaluate \metricName in two settings, an out-of-dataset (OOD) setting and an all-datasets (AD) setting.
    In the OOD setting, we train and tune \metricName on all datasets in \dataName \textit{except} the dataset we are evaluating on.
    This reflects the use case where we want to apply \metricName to evaluate a new dataset where we do not have human judgement scores.
    In the AD setting, we train on all datasets in \dataName and evaluate on all datasets.
    All results reported for \metricName are the average of three runs using the best set of hyperparameters found on the validation set of \dataName.
    
    \vspace{2mm}
    \noindent\textbf{Baselines:}
    We compare \metricName against BLEU-1 \citep{Papineni2001BleuAM}, ROUGE-L \citep{Lin2004ROUGEAP}, METEOR \citep{Banerjee2005METEORAA}, and BERTScore \citep{Zhang2019BERTScoreET}.
    We also compare \metricName against a BERT-base model fine-tuned on the sentence similarity task, STS-B \citep{Cer2017SemEval2017T1}.
    Results for BERT STS-B are the average of three runs using the best set of hyperparameters found on the validation set of STS-B.
    All baselines are agnostic to the passage and the question.

    \subsection{Correlation Results}
        We evaluate the baselines and OOD \metricName in Table \ref{tab:experiments:metric-correlation} using Pearson correlation.
        \metricName outperforms the baseline metrics despite being trained in a out-of-dataset situation.
        METEOR does surprisingly well despite relying on \textit{n}-gram overlap to do evaluation.
        Interestingly, the sentence similarity model does better than the baseline metrics while falling behind \metricName.
        
        We also study whether having human judgements for a particular dataset helps.
        We present results in Table \ref{tab:experiments:lerc-all} on the validation set of \dataName when \metricName is trained in an AD setting.
        Having human judgements for the target dataset  is always helpful.

    \begin{table}[tb]
        \begin{center}
        \begin{tabular}{lc}
        	\toprule
            \bf Dataset & \bf Dev \textit{r} \\
            \midrule
            NarrativeQA & 0.805 \\
            MCScript & 0.816 \\
            CosmosQA & 0.864 \\
            SocialIQA & 0.820 \\
            DROP & 0.796 \\
            Quoref & 0.794 \\
            \bottomrule
        \end{tabular}
        \end{center}
        \caption{Pearson correlation on the validation set of \dataName with \metricName trained on all constituent datasets.} 
        \label{tab:experiments:lerc-all}
    \end{table}
    
    \begin{table}[t!]
        \begin{center}
        \begin{tabular}{p{1.85 in} c}
        	\toprule
            \bf Ablation & \bf Avg. Dev \textit{r} \\\midrule
            Ref. Only & 0.081 \\
            Cand. Only & 0.093 \\
            Ref. \& Cand.  & 0.742 \\
            Ques. \& Ref. \& Cand. & 0.723 \\
            Pass. \& Ques. \& Ref. \& Cand. & 0.726\\
            \addlinespace\hline\addlinespace
            \metricName (with pre-training)  & 0.755 \\
            \bottomrule 
        \end{tabular}
        \end{center}
        \caption{Partial-input ablations of LERC trained in an out-of-dataset fashion.
        Results are Pearson correlation on the validation set, averaged across all constituent datasets.}
        \label{tab:experiments:lerc-ablation}
    \end{table}
    
    \begin{table}[h!t]
        \small
        \begin{center}
        \begin{tabular}{p{0.5in} p{2.2 in}}
        	\toprule
            \bf Error Source & \bf Example \\
            \midrule
            Passage Use \textit{(22.5\%)} & 
                \textbf{Passage:} Edward takes charge and the children develop and expand the farmstead, aided by the entrepreneurial spirit of the younger brother Humphrey. They are assisted by a gypsy boy, Pablo, who they rescue from a pitfall trap.\newline 
                \textbf{Q:} Who do the children rescue from a trap?\newline 
                \textbf{Ref:} Pablo \quad \textbf{Cand:} A gypsy kid\newline
                \textbf{Human Score:} 4.6 \quad \textbf{\metricName:} 1.0 \\\\
            Same Meaning \textit{(35\%)}&
                \textbf{Passage:} The story centres on the relationship between Mrs Kitty Warren and her daughter, Vivie. Mrs. Warren, a former prostitute. \newline 
                \textbf{Q:} What did Mrs. Warren previously do for work?\newline 
                \textbf{Ref:} Prostitution \newline
                \textbf{Cand:} She was an escort. \newline
                \textbf{Human Score:} 4.6 \quad \textbf{\metricName:} 1.06  \\\\
            Opposite Meaning \textit{(15\%)} &
                \textbf{Passage:} Sasha hated her neighbours dog as it barked all day and night so after going to the shop and buying poisonous slug pellets, Sasha gave the dog some pills. \newline 
                \textbf{Q:} How would you describe Sasha? \newline 
                \textbf{Ref:} mean \quad \textbf{Cand:} kind \newline
                \textbf{Human Score:} 1 \quad \textbf{\metricName:} 4.32 \\\\
            Other \textit{(27.5\%)} & 
                \textbf{Passage:} The train was slow and ambling, so much so that we were 2 hours late when we arrived in Montreal, missing our connection.\newline 
                \textbf{Q:} What might be true if the freight trains didn't cause a delay ? \newline 
                \textbf{Ref:} They wouldn't have missed their connection \newline
                \textbf{Cand:} they couldn't help noticing their connection \newline
                \textbf{Human Score:} 1 \quad \textbf{\metricName:}  4.2 \\
            \bottomrule
        \end{tabular}
        \end{center}
        \caption{Error analysis of \metricName. We take the 10 validation instances per \textit{generative} dataset (40 total) with the largest difference between the score assigned by \metricName and the score assigned by humans. We then group the highest error instances by the sources of the error.}
        \label{tab:experiments:error analysis}
    \end{table}

  \subsection{Error Analysis of \metricName}
        We gather the 10 validation instances per generative dataset (40 instances total) with the highest absolute difference between the human judgement score and \metricName score.
        We categorize the errors made by \metricName in Table \ref{tab:experiments:error analysis}.
        A large source of error is the inability to leverage the passage correctly as well as handling large lexical gaps between references and correctly paraphrased candidates.
        The ``Other'' category includes understanding semantic roles and misspellings of the reference.

    \begin{table*}[t!]
        \begin{center}
        \begin{tabular}{lccccc}
        \toprule
                \bf  Metric & \bf  NarrativeQA & \bf  MCScript & \bf CosmosQA & \bf SocialIQA & \bf Avg. \\
                BLEU-1      & 53 & 54 & 52 & 55 & 53.5  \\
                ROUGE-L     & 53 & 57 & 53 & 53 & 61.2 \\
                METEOR      & 60 & 62 & 57 & 53 & 54 \\
                BERTScore   & 70 & 58 & \textbf{74} & 62 & 66 \\
                \addlinespace \hline \addlinespace
                BERT STS-B  & 70.6 & 70 & 59.3 & 66.6 & 66.6 \\
                \addlinespace \hline \addlinespace
                \metricName  & \textbf{80} & \textbf{87.3} & 72.6 & \textbf{81.3} & \textbf{80.3} \\
                \bottomrule
        \end{tabular}
        \end{center}
    \caption{Results of \metricName (OOD setting) and baselines evaluated on minimal pairs. Numbers are accuracy values: given a minimal pair of candidates, what percent of the time does a metric prefer the better candidate.}
    \label{tab:experiments:minimal pair}
    \end{table*}
    
    \subsection{Ablation Results}
        We study five ablations of OOD LERC with results in Table \ref{tab:experiments:lerc-ablation}.
        All ablations do not involve any pre-training.
        When looking at ablations of \metricName, several interesting phenomena emerge.
        
        Pre-training is important with such a complex input structure.
        Removing pre-training while still using the passage and question as input hurts performance.
        Ablations of \metricName that do not use the passage but still have the reference and candidate as input only fall slightly behind the complete metric.
        One explanation is that current generative QA models may not generate many candidates that would require the metric to use the passage.
        Therefore, even the complete version of \metricName may have learned to ignore the passage.
        We explore this in the following section when conducting an error analysis of \metricName.
        
        As sanity checks for dataset biases, we also evaluate impoverished ablations that should not perform well: when the model has access only to the reference or to the candidate.  
        These ablations correlate quite poorly with human judgments. 
        The correlation is slightly positive for both, however, perhaps measuring the grammaticality of a candidate, or the difficulty of matching long references.
    
    \subsection{Minimal Pair Results}
        We now present results on the set of minimal pairs. 
        We use these minimal pairs to evaluate preference: given a minimal pair of candidates ($\mathbf{c_1}$, $\mathbf{c_2}$), what percentage of the time does a metric prefer the better candidate?
        For cases where a metric assigns the same score to both candidates, we give a half-point.
        
        Results are reported in terms of accuracy in Table \ref{tab:experiments:minimal pair}.
        \textit{N}-gram based metrics are close to random, which aligns with intuition because minimal pairs were created such that both candidates have a similar token overlap with the reference.
        The sentence similarity model does much better, likely because it generalizes beyond token overlap.
        Finally, \metricName (OOD setting) does the best, suggesting that while there is still room for improvement, the phenomena targeted by the minimal pairs is captured when evaluated  using preference.
        
   \subsection{\metricName vs BLEU}
        To understand the differences in behavior between \metricName and the popular BLEU metric, we collect the 10 validation instances per generative dataset with the highest absolute difference between the BLEU-1 and \metricName score.
        We categorize the source of the differences in Table \ref{tab:experiments:lerc-vs-bleu}.
        In about 90\% of the cases, the gap is due to BLEU scoring candidates too low (e.g. not capturing paraphrases).
        In the remaining cases, the gap is due to \metricName over-scoring the candidate, usually due to the reference and candidate being similar (e.g. both are numbers).
        
        \begin{table}[tb!]
            \small
            \begin{center}
            \begin{tabular}{p{0.7in} p{2 in}}
            	\toprule
                \bf Difference Source & \bf Examples \\
                \midrule
                BLEU under-scores paraphrases \textit{(92.5\%)} & 
                    \textbf{Passage:} Tracy took Jesse's students to the park. Jesse had an emergency and asked her to.\newline 
                    \textbf{Q:} How would Jesse feel afterwards?\newline 
                    \textbf{Ref:} grateful \quad \textbf{Cand:} thankful\newline
                    \textbf{\metricName:} 5.0 \quad \textbf{BLEU-1:} 0 \newline
                    \textbf{Human Score:} 5\\\\
                \metricName overly sensitive \textit{(7.5\%)}&
                    \textbf{Passage:} By 17, Norman is the best swordsman in all of England; by the age of 18, he has a large bounty on his head, and by the age of 19, he leads the largest band of thieves in all of England.\newline 
                    \textbf{Q:} What age was Norman when there was a bounty on his head?\newline 
                    \textbf{Ref:} 18 \quad \textbf{Cand:} 19 \newline
                    \textbf{\metricName:} 5.0 \quad \textbf{BLEU-1:} 0 \newline
                    \textbf{Human Score:} 1 \\\\
                \bottomrule
        \end{tabular}
        \end{center}
        \caption{Analysis of \metricName vs BLEU-1. We take the 10 validation instances per \textit{generative} dataset (40 total) with the largest difference between the score assigned by \metricName and the score assigned by BLEU-1. We then group these instances by the source of the difference.}
        \label{tab:experiments:lerc-vs-bleu}
    \end{table}
\section{Related Work}
	There has been a long history of developing evaluation metrics, which have generally fallen into one of three categories.
	The first consists of metrics that use some variant of \textit{n}-gram matching \citep{Papineni2001BleuAM,Lin2004ROUGEAP,Banerjee2005METEORAA}.
	They are easy to implement, but lack flexibility by focusing only on token overlap.
	The second cateogry of metrics eschew some of the aforementioned issues by calculating a \textit{softer} similarity score using \textit{embeddings} of tokens \citep{Clark2019SentenceMS,Zhang2019BERTScoreET}.
	However, it is unclear how to tailor them to question answering, where the passage and question should be assimilated.
	The final category consists of metrics learned end-to-end from human judgements \citep{Cui2018LearningTE,Sellam2020BLEURTLR}.
	These metrics are flexible in that they can be tuned to the specific evaluation setting but depend on a large corpus of human judgement scores to train on.
	We hope that the release of \dataName pushes the development of QA metrics that fall into this category.
    
	\dataName is directly inspired by the annual WMT Metrics Shared Task \citep{Machcek2014ResultsOT,Stanojevi2015ResultsOT,Bojar2016ResultsOT,Bojar2017ResultsOT,Ma2018ResultsOT,Ma2019ResultsOT}.
	Participants submit automatic translations and human judgement scores are collected for the submitted translations.
	The annotations collected as part of the WMT Metrics Shared Task have made it easy to evaluate and create new translation metrics~\cite{Popovic2015chrFCN,Ma2017BlendAN,Shimanaka2018RUSERU}.
	In a similar vein, SummEval is a recently released dataset that evaluates a number of evaluation metrics for summarization \citep{Fabbri2020SummEvalRS}.
\section{Conclusion}
    We present \dataName, a dataset of human judgement scores for training and evaluating generative reading comprehension metrics.
    Using \dataName, we train a learned metric, \metricName, that outperforms all existing metrics and is much more robust when evaluated on a set of minimal pairs.
    
    While we have demonstrated that \metricName is a better metric for evaluating generative reading comprehension than any existing metric, considerable work remains. 
    Error analysis reveals that there exist gaps in \metricName's ability to handle certain phenomena, such as correctly leveraging the passage.
    Future work involves collecting data to addresses weaknesses of \metricName.
    We also anticipate a continual cycle of generative RC model and dataset developments that will enable easier collection of more diverse and useful candidates.
    This in turn will allow better learned metrics, which can be used to evaluate ever more complex models.

\section*{Acknowledgements}
We would like to thank AI2 for the funding to collect \dataName.
We would also like to thank members of AI2 and UCI NLP for looking over early drafts of the paper.
This paper is based upon work sponsored by the DARPA MCS program under Contract No. N660011924033 with the United States Office Of Naval Research.

\bibliography{emnlp2020}
\bibliographystyle{acl_natbib}
\newpage \phantom{asdf} \newpage

\appendix
\section*{Appendix}

\section{Details on Training \metricName}
    Training of \metricName is broken into pre-training on multiple-choice QA datasets followed by fine-tuning on human judgement scores.
    
    During pre-training, we used batch size of 32 and train for 4 epochs. 
    We tune the learning rate (\{1e-5, 2e-5, 3e-5\}) over held out questions using a single runs' loss.
    We use accuracy as the criteria to pick the best pre-trained model.
    
    We then take the best pre-trained model and fine-tune on human judgement scores in \dataName.
    We again fix the batch size at 32 and train for 3 epochs, tuning the learning rate (\{1e-5, 2e-5, 3e-5\}) over the validation set of \dataName using the average of three runs.
    We use Pearson correlation to pick the best fine-tuned model.
    When LERC is trained in an OOD setting, we do not tune on the held-out dataset.

\section{Details on Baselines}
    We use implementations of BLEU, METEOR, and ROUGE using Microsoft MS COCO evaluation scripts \footnote{https://github.com/salaniz/pycocoevalcap}.
    We removed question marks, periods, and exclamation marks from references and candidates when evaluating with BLEU, METEOR, and ROUGE.
    
    The hash-code for BERTScore is \path{roberta-large_L17_no-idf_version=0.3.6(hug_trans=3.0.2)}.
    
    We fine-tune BERT-base on STS-B as another baseline.
    We use a batch size of 32 and train for 4 epochs.
    We tune the learning rate (\{1e-5, 2e-5, 3e-5\}) over the validation set of STS-B using the average of three runs.

\section{Computational Resources}
    All experiments on conducted on a NVIDIA Titan RTX with 24 GB of RAM.
    Pre-training of \metricName takes about 3.5 hours while fine-tuning (one run) takes roughly 20 minutes.

\section{Details on Mechanical Turk}
\label{sec:appendix_mechanical_turk}
    Collecting \dataName involves three stages: a qualification testing stage, a trial stage, and the full dataset collection stage.
    
    During qualification testing, workers are given 10 candidates to label, and they must score 80\% to pass the test.
    After qualification testing, we run a small trial.
    During this trial, we release 200 candidates and gather 5 human judgements per candidate to get a sense of annotation agreement and to see if our instructions and examples need to be revised.
    Finally, during the full dataset collection process we solicit human judgements on all candidates.
    Here, each HIT is an aggregate of 10 candidates that all share the same passage to amortize the cost of reading the passage and workers are paid 40 cents per HIT.\footnote{This amount is set by the authors manually working on this task. We estimate that it takes between a minute and a half to two and a half minutes to complete a HIT depending on the dataset.} 
    During dataset collection, we randomly sample annotations to check for quality and remove workers that consistently do a poor job.
    
    Workers are paid for working on any of the three stages. The total cost of collecting \dataName is about \$6,000.

\section{Correlation Results based on Generation Source}
	We supplement Table \ref{tab:experiments:metric-correlation} by calculating correlation results per generation source for the generative datasets in Table \ref{tab:appendix:generation-correlation}.
	We find that \metricName handles candidates from different generation sources with roughly the same performance.

    \begin{table}[t!]
        \begin{center}
        \begin{tabular}{p{1.9 in} c}
            \toprule
            \bf Dataset/Generation Source & \bf Avg. Dev \textit{r} \\\midrule
            CosmosQA &  \\
 				\hspace{6pt} \small \textit{Backtranslation} & 0.714 \\
            	\hspace{6pt} \small \textit{GPT-2} & 0.636 \\
			MCScript &  \\
            	\hspace{6pt} \small \textit{Backtranslation} & 0.545 \\
            	\hspace{6pt} \small \textit{GPT-2} & 0.661 \\
            	\hspace{6pt} \small \textit{MHPG} & 0.742 \\
			NarrativeQA &  \\
            	\hspace{6pt} \small \textit{Backtranslation} & 0.707 \\
            	\hspace{6pt} \small \textit{GPT-2} & 0.791 \\
            	\hspace{6pt} \small \textit{MHPG} & 0.814 \\
			SocialIQA &  \\
            	\hspace{6pt} \small \textit{Backtranslation} & 0.602 \\
            	\hspace{6pt} \small \textit{GPT-2} & 0.596 \\
            \bottomrule 
        \end{tabular}
        \end{center}
        \caption{Correlation on the validation set (OOD setting) broken down by the source of the generation.}
        \label{tab:appendix:generation-correlation}
    \end{table}
\end{document}